\documentclass[runningheads]{llncs}
\usepackage{graphicx}
\usepackage{amsmath,amssymb} % define this before the line numbering.
\usepackage{color}
\usepackage{times}
\usepackage{epsfig}
\usepackage{amsmath}
\usepackage{amssymb}
\usepackage{makeidx}         % allows index generation
\usepackage{microtype}
\usepackage{array}
\usepackage{hhline}
\usepackage{multirow}
\usepackage{wrapfig}
\usepackage{dsfont}
\usepackage{caption}
\usepackage{enumitem}
\usepackage[width=122mm,left=12mm,paperwidth=146mm,height=193mm,top=12mm,paperheight=217mm]{geometry}

\newcommand{\myparagraph}[1]{\smallskip\noindent\textbf{#1.}}

\begin{document}
% \linenumbers
\pagestyle{headings}
\mainmatter

\title{A Unified Framework for Multi-View Multi-Class Object Pose Estimation} 

\titlerunning{Multi-View Multi-Class Object Pose Estimation}
% \def\ECCV18SubNumber{1848}  % Insert your submission number here
% \titlerunning{ECCV-18 submission ID \ECCV18SubNumber}
% \authorrunning{ECCV-18 submission ID \ECCV18SubNumber}
% \author{Anonymous ECCV submission}
% \institute{Paper ID \ECCV18SubNumber}
%
% \author{Chi Li\orcidID{0000-0002-5957-5680} \and Jin Bai\orcidID{0000-0002-0653-0542} \and Gregory D. Hager\orcidID{0000-0002-6662-9763}}
\author{Chi Li \and Jin Bai \and Gregory D. Hager}

%
%Please write out author names in full in the paper, i.e. full given and family names.
%If any authors have names that can be parsed into FirstName LastName in multiple ways, please include the correct parsing, in a comment to the volume editors:
\index{Hager, Gregory}
%(Do not uncomment it, because you may introduce extra index items if you do that, we will use scripts for introducing index entries...)
\authorrunning{C. Li, J. Bai and G. Hager}
% Replace with shorter version of the author list. If there are more authors than fits a line, please use A. Author et al.
%
\institute{Department of Computer Science, Johns Hopkins University \\
\email{\{chi\_li,jbai12,hager\}@jhu.edu}}

\maketitle

\begin{abstract}
One core challenge in object pose estimation is to ensure accurate and robust performance for large numbers of diverse foreground objects amidst complex background clutter.
In this work, we present a scalable framework for accurately inferring six Degree-of-Freedom (6-DoF) pose for a large number of object classes from single or multiple views.
To learn discriminative pose features, we integrate three new capabilities into a deep Convolutional Neural Network (CNN): an inference scheme that combines both classification and pose regression based on a uniform tessellation of the Special Euclidean group in three dimensions (SE(3)), the fusion of class priors into the training process via a tiled class map, and an additional regularization using deep supervision with an object mask.
Further, an efficient multi-view framework is formulated to address single-view ambiguity. 
We show that this framework consistently improves the performance of the single-view network.
We evaluate our method on three large-scale benchmarks: YCB-Video, JHUScene-50 and ObjectNet-3D. 
Our approach achieves competitive or superior performance over the current state-of-the-art methods.
\keywords{Object pose estimation, multi-view recognition, deep learning}
\end{abstract}

\section{Introduction}

\begin{figure}[t]
  \centering
    \includegraphics[width=1.0\linewidth]{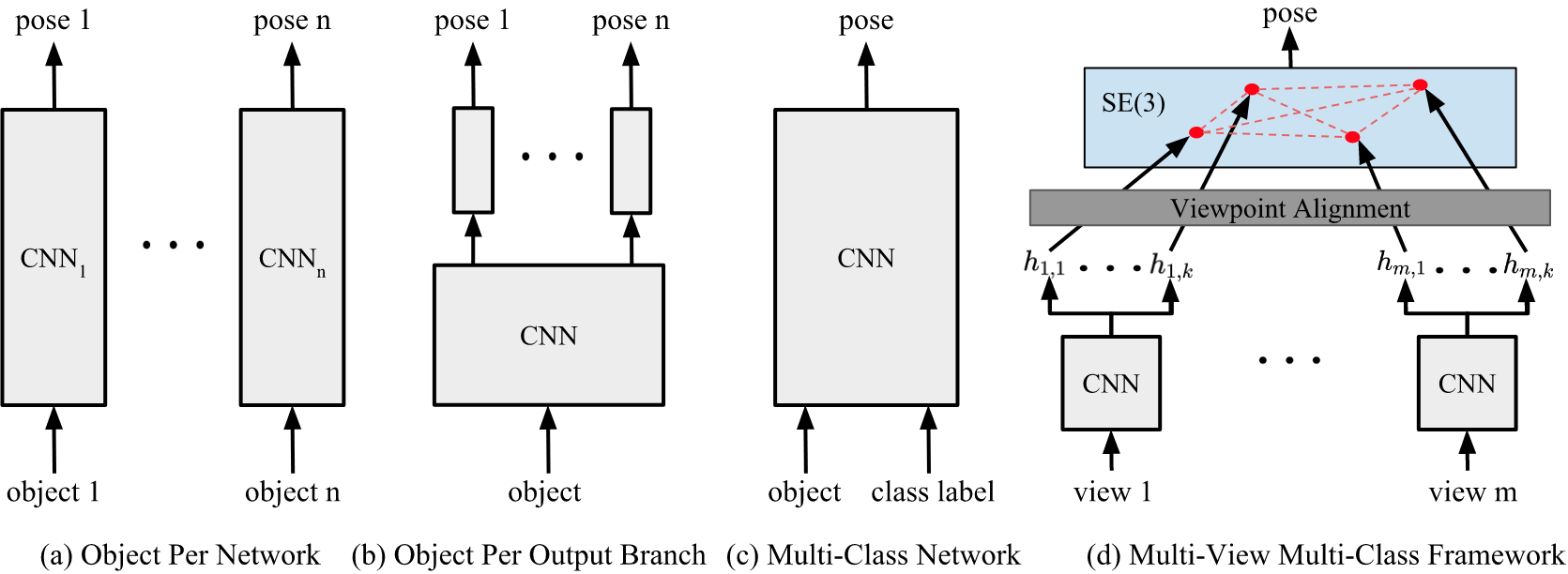}
  \caption{Illustration of different learning architectures for single-view object pose estimation: (a) each object is trained on an independent network; (b) each object is associated with one output branch of a common CNN root; and (c) our network with single output stream via class prior fusion. Figure (d) illustrates our multi-view, multi-class pose estimation framework where $h_{m,k}$, the $k$-th pose hypothesis on view $m$, is first aligned to a canonical coordinate system and then matched against other hypotheses for pose voting and selection.}
 \label{fig:mvmc_frame}
\end{figure}

% overview of pose estimation, why is it important? 
Estimating 6-DoF object pose from images is a core problem for a wide range of applications including robotic manipulation, navigation, augmented reality and autonomous driving. While numerous methods appear in the literature \cite{linemod-accv12,wohlhart2015learning,balntas2017pose,tjaden2017real,brachmann2014learning,doumanoglou2016recovering,kehl2016deep,michel2017global}, scalability (to large numbers of objects) and accuracy continue to be critical issues that limit existing methods.
% Traditional template-based methods compute object pose by matching image observations to object templates sampled from a constrained viewing sphere~\cite{linemod-accv12,wohlhart2015learning,balntas2017pose,tjaden2017real}.
% In addition, bottom-up approaches classify image pixels into semantic classes and instance parts via local features, followed by pose hypothesis refinement~\cite{brachmann2014learning,doumanoglou2016recovering,kehl2016deep,michel2017global}. 
% However, these two work categories rely on models whose complexity dramatically grows with the number of object instances, and generalize poorly to unseen viewpoints and object instances.
% Moreover, single-view ambiguity and occlusion in cluttered scenes are fundamental bottlenecks that prevent these methods from predicting highly accurate pose. 
Recent work has attempted to leverage the power of deep CNNs to surmount these limitations~\cite{su2015render,massa2016crafting,xiang2016objectnet3d,mousavian20173d,tekin2017real,kehl2017ssd,xiang2017posecnn,rad2017bb8}.
One naive approach is to train a network to estimate the pose of each object of interest (Fig.~\ref{fig:mvmc_frame} (a)). More recent approaches follow the principle of ``object per output branch'' (Fig.~\ref{fig:mvmc_frame} (b)) whereby each object class\footnote{An object class may refer to either an object instance or an object category.} is associated with an output stream connected to a shared feature basis~\cite{xiang2017posecnn,kehl2017ssd,su2015render,massa2016crafting,rad2017bb8}. 
In both cases, the size of the network increases with the number of objects, which implies that large amounts of data are needed for each class to avoid overfitting. 
In this work, we present a multi-class pose estimation architecture (Fig.~\ref{fig:mvmc_frame} (c)) which receives object images and class labels provided by a detection system and which has a single branch for pose prediction.
As a result, our model is readily scalable to large numbers of object categories and works for unseen instances while providing robust and accurate pose prediction for each object. % by fusing class prior into feature development.

The ambiguity of object appearance and occlusion in cluttered scenes is another problem that limits the application of pose estimation in practice.
One solution is to exploit additional views of the same instance to compensate for recognition failure from a single view. 
However, naive ``averaging'' of multiple single-view pose estimates in SE(3)~\cite{chirikjian2018pose} does not work due to its sensitivity to incorrect predictions.
Additionally, most current approaches to multi-view 6-DoF pose estimation~\cite{slam++13,chi-icra16,erkent2016integration} do not address single-view ambiguities caused by object symmetry.
This exacerbates the complexity of view fusion when multiple correct estimates from single views do not agree on SE(3).
Motivated by these challenges, we demonstrate a new multi-view framework (Fig.~\ref{fig:mvmc_frame} (d)) which selects pose hypotheses, computed from our single-view multi-class network, based on a distance metric robust to object symmetry.

In summary, we make following contributions to scalable and accurate pose estimation on multiple classes and multiple views:
\begin{itemize}
\item We develop a multi-class CNN architecture for accurate pose estimation with three novel features: a) a single pose prediction branch which is coupled with a discriminative pose representation in SE(3) and is shared by multiple classes; b) a method to embed object class labels into the learning process by concatenating a tiled class map with convolutional layers; and c) deep supervision with an object mask which improves the generalization from synthetic data to real images.
\item We present a multi-view fusion framework which reduces single-view ambiguity based a voting scheme. An efficient implementation is proposed to enable fast hypothesis selection during inference. 
% We empirically validate that our multi-view algorithm consistently improves the single-view pose estimation performance.
\item 
We show that our method provides state-of-the-art performance on public benchmarks including YCB-Video~\cite{xiang2017posecnn}, JHUScene-50~\cite{chi-icra16} for 6-DoF object pose estimation~\cite{xiang2017posecnn,chi-icra16}, and ObjectNet-3D for large-scale viewpoint estimation~\cite{xiang2016objectnet3d}.
Further, we present a detailed ablative study on all benchmarks to empirically validate the three innovations in the single-view pose estimation network.
\end{itemize}
% paper structure
% In the remainder of this paper, we review related work in Sec.~\ref{sec:related}.
% The multi-class single-view network is introduced in Sec.~\ref{sec:mcn} and the multi-view framework is presented in Sec.~\ref{sec:mv}.
% We evaluate our method in Sec.~\ref{sec:exp} and conclude the paper in Sec.~\ref{sec:conclude}.

\section{Related Work}
\label{sec:related}
We first review three categories of work on single-view pose estimation and then investigate recent progress on multi-view object recognition. 

\textbf{Template Matching.}
Traditional template-based methods compute 6-DoF pose of an object by matching image observations to hundreds or thousands of object templates that are sampled from a constrained viewing sphere~\cite{linemod-accv12,wohlhart2015learning,balntas2017pose,tjaden2017real}.
% LINEMOD system~\cite{linemod-accv12} proposes object templates as cropped object images associated with viewpoints that are uniformly sampled on a constrained range of viewing sphere.
% To compute 6-DoF pose of an object, hundreds or thousands of object templates are matched to region proposals provided from a detection system~\cite{linemod-accv12,tjaden2017real}.
Recent approaches apply deep CNNs as end-to-end matching machines to improve the robustness of template matching ~\cite{wohlhart2015learning,balntas2017pose,krull2015learning}.
Unfortunately, these methods do not scale well in general because the inference time grows linearly with the number of objects.
Moreover, they generalize poorly to unseen object instances as shown in \cite{balntas2017pose} and suffer from poor domain shift from synthetic to real images.  

\textbf{Bottom-Up Approaches.}
% With the emergence of commercial RGB-D sensors, 3D point cloud becomes readily available in real-time analysis.
Given object CAD models, 6-DoF object pose can be inferred by registering a CAD model to part of a scene 
using coarse-to-fine ICP~\cite{zeng2017multi}, Hough voting~\cite{tejani2014latent}, RANSAC~\cite{objransac-accv10} and heuristic 3D descriptors~\cite{Tombari-icip11,fpfh}.
More principled approaches use random forests to infer local object coordinates for each image pixel based on hand-crafted features~\cite{Brachmann-eccv14,brachmann2016uncertainty,michel2017global} or auto-encoders~\cite{doumanoglou2016recovering,kehl2016deep}. 
% Subsequently, energy-based global optimization is used to estimate and refine object poses of multiple instances~\cite{Brachmann-eccv14,michel2017global}.
However, local image patterns are ambiguous for objects with similar appearance, which prevents this line of work from being applied to generic objects and unconstrained background clutter.

\textbf{Learning End-to-End Pose Machines.}
This class of work deploys deep CNNs to learn an end-to-end mapping from a single RGB or RGB-D image to object pose. 
\cite{su2015render,massa2016crafting,mousavian20173d,xiang2016objectnet3d} train CNNs to directly predict the Euler angles of object instances and then apply them to unseen instances from the same object categories.
% The main objective of these methods is to recognize object viewpoints from an unconstrained cluttered scenes and generalize to unseen instances of an object category that is trained before.
Other methods decouple 6-DoF pose into rotation and translation components and infer each independently.
% \cite{tekin2017real,rad2017bb8} directly regresses 2D locations of projected bounding box corners and recovers 3D pose from 2D projections via PnP algorithm~\cite{lepetit2009epnp}.
SSD-6D~\cite{kehl2017ssd} classifies an input into discrete bins of Euler angles and subsequently estimates 3D position by fitting 2D projections to a detected bounding box.
PoseCNN~\cite{xiang2017posecnn} regresses rotation with a loss function that is robust to object symmetry, and follows this with a bottom-up approach to vote for the 3D location of the object center via RANSAC.
In contrast to the above, our method formulates a discriminative representation of 6-DoF pose that enables predictions of both rotation and translation by a single forward pass of a CNN, while being scalable to hundreds of object categories. 
% Our method formulates a generic and discriminative representation of 6-DoF pose which enables direct prediction of object rotation and translation from either RGB or RGB-D data.
% Moreover, our approach can be directly applied in unconstrained environment for recognizing viewpoints of unseen instances, in the scale of hundreds of object categories.

\textbf{Multi-View Recognition.}
In recent years, several multi-view systems have been developed to enhance 3D model classification~\cite{su2015multi,johns2016pairwise}, 2D object detection~\cite{lai2012detection,pillai2015monocular} and semantic segmentation~\cite{chi-iros16,tateno2017cnn,zeng2017multi}.
For 6-DoF pose estimation, SLAM++~\cite{slam++13} is an early representative of a multi-view pose framework which jointly optimizes poses of both the detected object and the cameras. 
\cite{chi-iros16} computes object pose by registering 3D object models over an incrementally reconstructed scene via a dense SLAM system. 
These two methods are difficult to scale because they rely on ~\cite{objransac-accv10} whose running time grows linearly to the number of objects.
A more recent method~\cite{erkent2016integration} formulates a probabilistic framework to fuse pose estimates from different views.
However, it requires computation of marginal probability over all subsets of a given number of views, which is computationally prohibitive when the number of views and/or objects is large.

\section{Single-View Multi-Class Pose Estimation Network}
\label{sec:mcn}
\begin{figure}[t]
  \centering
    \includegraphics[width=1.0\linewidth]{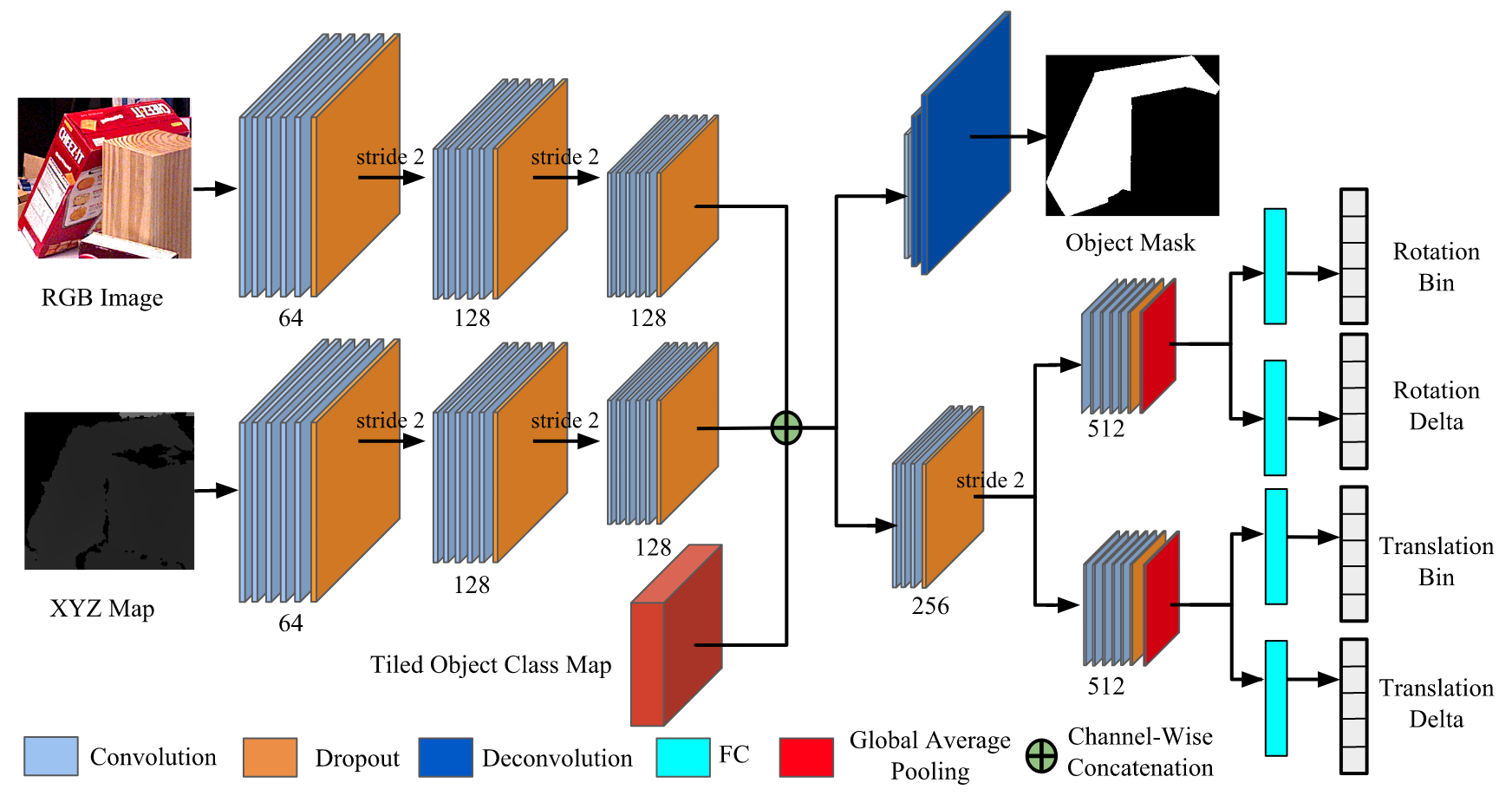}
  \caption{Multi-class network architecture for a single view; the figure shows the actual number of layers used in our implementation. We note that the XYZ map which represents normalized 3D coordinates of each image pixel. If depth data is not available, this stream is omitted.  }
 \label{fig:mv_network}
\end{figure}

In this section, we introduce a CNN-based architecture for multi-class pose estimation (Fig. \ref{fig:mv_network}). 
The input can be an RGB or RGB-D image region of interest (ROI) of an object provided by arbitrary object detection algorithm.  
The network outputs represent both the rotation $R$ and the translation $T$ of a 6-DoF pose $(R, T)$ in SE(3). 

We first note that the a single rotation $R$ relative to the camera corresponds to different object appearances in image domain when $T$ varies.
This issue has been discussed in \cite{mousavian20173d} in the case of 1-D yaw angle estimation. 
To create a consistent mapping from the ROI appearance to $(R, T)$, we initially rectify the annotated pose to align to the current viewpoint as follows. 
We first compute the 3D orientation $\vec{v}$ towards the center of the ROI $(x, y)$: $\vec{v} = [(x - c_x) / f_x, (y - c_y) / f_y, 1]$, where $(c_x, c_y)$ is the 2D camera center and $f_x, f_y$ are the focal lengths for $X$ and $Y$ axes.
Subsequently, we compute rectified XYZ axes $[X_{\vec{v}}, Y_{\vec{v}}, Z_{\vec{v}}]$ by aligning the Z axis $[0, 0, 1]$ to $\vec{v}$. 
\begin{equation}
X_{\vec{v}} = [0, 1, 0] \times Z_{\vec{v}}, \ Y_{\vec{v}} = Z_{\vec{v}} \times X_{\vec{v}}, \ Z_{\vec{v}} = \frac{\vec{v}}{\|\vec{v}\|_2} 
\end{equation}
where symbol $\times$ indicates the cross product of two vectors.
Finally, we project $(R, T)$ onto $[X_{\vec{v}}, Y_{\vec{v}}, Z_{\vec{v}}]$ and obtain the rectified pose $(\widetilde{R}, \widetilde{T})$: $\widetilde{R} = R_{\vec{v}} \cdot R$ and $\widetilde{T} = R_{\vec{v}} \cdot T$, where $R_{\vec{v}} = [X_{\vec{v}};Y_{\vec{v}};Z_{\vec{v}}]$. 
We refer readers to the supplementary material for more details about the rectification step.
When depth is available, we rectify the XYZ value of each pixel by $R_{\vec{v}}$ and construct a normalized XYZ map by centering the point cloud to the median along each axis.

Figure~\ref{fig:mv_network} illustrates the details of our network design.
Two streams of convolutional layers receive RGB image
and XYZ map respectively and the final outputs are bin and delta vectors (described below) for both rotation and translation (Sec.~\ref{sec:bd}).
These two streams are further merged with class priors (Sec.\ref{sec:fuse_class}) and deeply supervised by object mask  (Sec.~\ref{sec:ds_mask}).
When depth data is not available, we simply remove the XYZ stream.

\subsection{Bin \& Delta Representation for SE(3)}
\label{sec:bd}

Direct regression to object rotation $R$ has been shown to be inferior to a classification scheme over discretized SO(3)\footnote{SO(3) is the Special Orthogonal group of rations in three dimensions}~\cite{ren2015faster,mousavian20173d,kehl2017ssd}.
% In this section, we present a new pose representation that is used for the training and inference of our multi-class pose network. 
% Image-based pose estimation is essentially a complex multi-modal regression problem due to the single-view ambiguity, object symmetry, occlusion and background clutter.
% As discussed in \cite{ren2015faster,mousavian20173d,kehl2017ssd}, direct regression using L2 loss is inferior to splitting the regression space into different subspaces (or bins) and regressing the deviation (or delta) to the groundtruth within each subspace.
% The hidden assumption is that each individual subspace only covers a single mode in a multi-modal distribution. 
% For example, Faster R-CNN~\cite{ren2015faster} manually defines the anchor boxes as the bin centers for the space of bounding boxes and \cite{mousavian20173d} creates bins in 1-D yaw angle space via uniform gridding.
One common discretization of SO(3) is to bin along each Euler angle $(\alpha, \beta, \gamma)$ (i.e. yaw, pitch and roll)~\cite{su2015render,kehl2017ssd}. 
However, this binning scheme yields a non-uniform tessellation of SO(3).
Consequently, a small error on one Euler angle may be magnified and result in a large deviation in the final rotation estimate.
% To see this, consider a rotation $R=R_\gamma R_\beta R_\alpha$ composed by a sequence of rotations based on its Euler angles. 
% The small error $\delta$ in predicting $\alpha$ can lead to large error in final prediction: $d(R_\gamma R_\beta R_\alpha, R_\gamma R_\beta R_{\alpha+\delta})$, where $d(R_1, R_2) = \frac{1}{2}\|\log(R_1^TR_2)\|_F$ is the geodesic distance between two rotations $R_1$ and $R_2$.
% In other words, the neighborhood of each bin center is not smooth and highly non-linear.
%, which is not appropriate for bin classification and delta regression. 
In the following, we formulate two new bin \& delta representations which uniformly partition both SO(3) and R(3). 
They are further coupled with a classification \& regression scheme for learning discriminative pose features.
% These two representations jointly describe SE(3).
% In addition, we detail the scoring scheme of such representations for training.

\noindent
\textbf{Almost Uniform Partition of SO(3).}
\label{sec:bd_so(3)}
We first exploit the sampling technique developed by \cite{yan2012almost} to generate $N$ rotations $\{\hat{R}_1,...,\hat{R}_N\}$ that are uniformly distributed on SO(3). 
These $N$ rotations are used as the centers of $N$ rotation bins in SO(3). These are shared between different object classes. 
Given an arbitrary rotation matrix $R$, we convert it to a bin and delta pair $(\vec{b}^R,\vec{d}^R)$ based on $\{\hat{R}_1,...,\hat{R}_N\}$.
The bin vector $\vec{b}^R$ contains $N$ dimensions where the $i$-th dimension $\vec{b}^R_i$ indicates the confidence of $R$ belonging to bin $i$. 
$\vec{d}^R$ stores $N$ rotations (i.e. quaternions in our implementation) where the $i$-th rotation $\vec{d}^R_i$ is the deviation from $\hat{R}_i$ to $R$.
During inference, we take the bin with maximum score and apply the corresponding delta value to the bin center to compute the final prediction. 
In training, we enforce a sparse confidence scoring scheme for $(\vec{b}^R,\vec{d}^R)$ to supervise the network:
% Given a rotation $R$, we only activate some representative bins and deltas:
\begin{equation}
\vec{b}^R_i = \begin{cases}
        \theta_1 & : i \in NN_1(R) \\
        \theta_2 & : i \in NN_k(R) \setminus NN_1(R) \\
        0  & : \text{Otherwise} \\
     \end{cases}, \ \ \
\vec{d}^R_i = \begin{cases}
        R\cdot\hat{R}^T_i & : i \in NN_k(R) \\
        0  & : \text{Otherwise} \\
\end{cases}
\label{eq:conf_score}
\end{equation}
where $\theta_1 \gg \theta_2$ and $NN_k(R)$ is the set of $k$ nearest neighbors of $R$ among $\{\hat{R}_1,...,\hat{R}_N\}$ in terms of the geodesic distance $d(R_1, R_2) = \frac{1}{2}\|\log(R_1^TR_2)\|_F$ between two rotations $R_1$ and $R_2$.
Note that we design delta $\vec{d}_i$ to achieve $R = \vec{d}^R_i\cdot \hat{R}_i$ and not $R = \hat{R}_i \cdot \vec{d}^R_i$ because the former is numerically more stable.
% For the second case, small prediction error on $\vec{d}^R_i$ may cause large error on final prediction of $R$ even if the bin prediction is correct.
Specifically, if $d$ is the prediction of $d^R_i$ with error $\delta$ such that $d = \delta \cdot d^R_i$, the error of final prediction $R'$ is also $\delta$ because $R' = d \cdot \hat{R}_i = \delta R$. 
If we define $R=\hat{R}_i\cdot d^R_i$ instead, then $R' = \hat{R}_i \cdot d = (\hat{R}_i \delta (\hat{R}_i)^{-1}) R$ and the error will be $\hat{R}_i \delta (\hat{R}_i)^{-1}$.
Thus, the $\delta$ error of $d^R_i$ may be magnified in the final rotation estimate $R$.

\noindent
\textbf{Gridding XYZ Axes.}
\label{sec:bd_tran}
The translation vector is the 3D vector from the camera origin to the object center.
To divide the translation space, we uniformly grid X, Y and Z axes independently.
For RGB images, we align the X and Y axes to image coordinates and the Z axis is optical axis of the camera.  
% Therefore, X and Y coordinates indicate the image location of projected 3D object center and Z value remains as the depth distance. 
We also rescale the ROI to a fixed scale for the CNN, so we further adjust the $Z$ value of each pixel to $Z'$ such that image scale is consistent to the depth value: $Z' = Z \cdot \frac{s'}{s}$, where $s'$ and $s$ are image scales before and after rescaling, respectively.
When depth data is available, the XYZ axes are simply chosen to be the coordinate axes of normalized point cloud.

We now discuss how to construct the bin \& delta pair $(\vec{b}^{T_x}, \vec{d}^{T_x})$ for X axis; the 
Y and Z axes are done in the same way.
We first create $M$ non-overlapping bins of equal size $\frac{s_{max} - s_{min}}{M}$ between $[s_{min}, s_{max}]$ \footnote{$s_{min}$ and $s_{max}$ may vary across different axes}.
When the X value is lower than $s_{min}$ (or larger than $s_{max}$), we assign it to the first (or last bin).
During inference, we compute the X value by adding the delta to the bin center which has the maximum confidence score. 
During training, similar to Eq.~\ref{eq:conf_score}, we compute $\vec{b}^{T_x}$ of an X value by finding its $K'$ nearest neighbors among $M$ bins.
Then, we assign $\theta'_1$ for the top nearest neighbor and $\theta'_2$ for the remaining $K-1$ neighbors ($\theta'_1 \gg \theta'_2$).
Correspondingly, the delta values of the $K'$ nearest neighbor bins are deviations from the bin centers to the actual X value and others are $0$.
Finally, we concatenate all bins and deltas of X, Y and Z axes: $\vec{b}^T=[\vec{b}^{T_x},\vec{b}^{T_y},\vec{b}^{T_z}]$ and $\vec{d}^T=[\vec{d}^{T_x},\vec{d}^{T_y},\vec{d}^{T_z}]$.
One alternative way of dividing translation space is to apply joint griding over XYZ space.
However, the total number of bins grows exponentially as $M$ increases and we found no performance gain by doing so in practice.

\subsection{Fusion of Class Prior}
\label{sec:fuse_class}
Many existing methods assume known object class labels, provided by a detection system, prior to pose analysis~\cite{xiang2017posecnn,kehl2017ssd,ren2015faster,massa2016crafting,balntas2017pose}.
However, they ignore the class prior during training and only apply it during inference.
Our idea is to directly incorporate this known class label into the learning process of convolutional filters for pose.
This is partly inspired by prior work on CNN-based hand-eye coordination learning~\cite{levine2016learning} where a tiled robot motor motion map is concatenated with one hidden convolutional layer for predicting the grasp success probability.
Given the class label of the ROI, we create a one-hot vector where the entry corresponding to the class label is set to $1$ and all others to $0$.
We further spatially tile this one-hot vector to form a 3D tensor with size $H\times W\times C$, where $C$ is the number of object classes and $H,W$ are height and width of a convolutional feature map at an intermediate layer chosen as part of the network design.
As shown in Fig.~\ref{fig:mv_network}, we concatenate this tiled class tensor with the last convolutional layers of both color and depth streams along the filter channel.
Therefore, the original feature map is embedded with class labels at all spatial locations and the subsequent layers are able to model class-specific patterns for pose estimation.
This is critical in teaching the network to develop compact class-specific filters for each individual object while taking advantage of a shared basis of low level features for robustness.

\subsection{Deep Supervision with Object Segmentation}
\label{sec:ds_mask}

Due to limited availability of pose annotations on real images, synthetic CAD renderings are commonly used as training data for learning-based pose estimation methods~\cite{xiang2017posecnn,linemod-accv12,kehl2017ssd}.
We take this approach but, following \cite{chi-cvpr17}, we also incorporate the deep supervision of an object mask at a hidden layer, (shown in Fig.~\ref{fig:mv_network}) for additional regularization of the training process. We can view the object mask as an intermediate result for the final task of 6-DoF pose estimation.
That is, good object segmentation is a prerequisite for the final success of pose estimation.
Moreover, a precisely predicted object mask benefits a post-refinement step such as Iterative Closest Point (ICP).

% As such, we enforce an inference sequence of first predicting object segmentation and then inferring object pose. 
To incorporate the mask with the feature and class maps (Sec.~\ref{sec:fuse_class}), we append one output branch for the object mask which contains one convolutional layer followed by two de-convolution layers with upsampling ratio $2$.
We assume that the object of interest  dominates the input image so that only a binary mask (``1'' indicates object pixel and ``0'' means background or other objects) is needed as an auxiliary cue.
As such, the size of the output layer for binary segmentation prediction is fixed regardless of the number of object instances in database, which enables our method to scale well to large numbers of objects. 
Conversely, when multiple objects appear in a scene, we must rely on some detection system to  ``roughly'' localize them in the 2D image first.
% As shown in our experiment (Sec.~\ref{sec:exp_scene50}), this deep supervision design is particularly helpful in generalizing from synthetic to real data.

\subsection{Network Architecture}

The complete loss function for training the network consists of five loss components over the segmentation map, the rotation, and the three translation components:
\begin{equation}
\mathcal{L} = l_{seg} + l_{R_b}(\widetilde{\vec{b}^R}, \vec{b}^R) + l_{R_d}(\widetilde{\vec{d}^R}, \vec{d}^R) + \sum_{i\in\{X,Y,Z\}} \Big(l_{T_b}(\widetilde{\vec{b}^{T_i}}, \vec{b}^{T_i}) + l_{T_d}(\widetilde{\vec{d}^{T_i}}, \vec{d}^{T_i})\Big)
\end{equation}
where $\widetilde{\vec{b}^R}$, $\widetilde{\vec{d}^R}$, $\widetilde{\vec{b}^{T_i}}$ and $\widetilde{\vec{d}^{T_i}}$ are the bin and delta estimates of the groundtruth $\vec{b}^R$, $\vec{d}^R$, $\vec{b}^{T_i}$ and $\vec{d}^{T_i}$, respectively.
% $l_{seg}$ is the standard segmentation loss where we apply softmax cross entropy loss over each pixel location.
% $l_{seg}$ indicates the loss for segmentation outputs (Sec.\ref{sec:ds_mask}).
We apply cross-entropy softmax to segmentation loss $l_{seg}$ on each pixel location and to the bin losses $l_{R_b}$ and $l_{T_b}$.
We employ L2 losses for the delta values $l_{R_d}$ and $l_{T_d}$.
All losses are simultaneously backpropagated to the network to update network parameters on each batch.
For simplicity, we apply loss weight $1$ for each loss term. 

Each convolutional layer is coupled with a batch-norm layer~\cite{ioffe2015batch} and ReLU.
The size of all convolutional filters is $3$x$3$.
The output layer for each bin and delta is constructed with one global average pooling (GAP) layer followed by one fully connected (FC) layer with $512$ neurons.
We employ a dropout~\cite{krizhevsky2012imagenet} layer before each downsampling of convolution with stride $2$. 
We deploy $23$ layers in total.

\section{Multi-View Pose Framework}
\label{sec:mv}

In this section, we present a multi-view framework which refines the outputs of our single-view network (Sec.~\ref{sec:mcn}) during the inference stage.
We assume that camera pose of each frame in a sequence is known.
In practice, camera poses can be provided by many SLAM systems such as Kinect Fusion~\cite{kinectfusion}. 

\subsection{Motivation}

Recall that we can obtain top-$K$ estimates from all subspaces in SE(3) including SO(3), X, Y, and Z spaces (Sec.~\ref{sec:bd}).
Therefore, we can compute $K^4$ pose hypotheses 
\begin{wrapfigure}{r}{0.5\textwidth}
	\centering
    \includegraphics[width=1.0\linewidth]{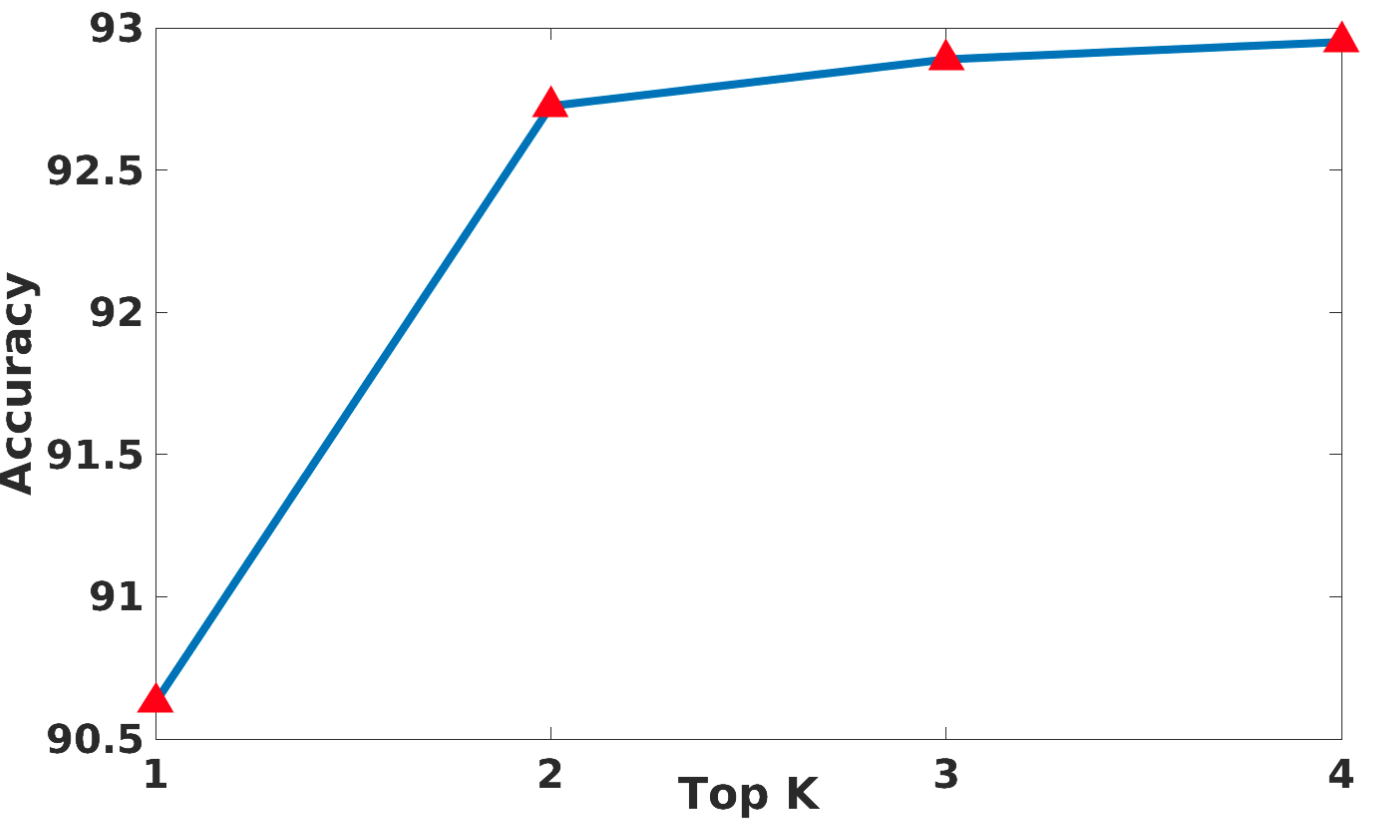}
  \caption{Top-$K$ accuracies of our single-view pose network on YCB-Video~\cite{xiang2017posecnn}. }
 \label{fig:top_k}
% \end{center}
\end{wrapfigure}
by composing top-k results from all subspaces.
In turn, we compute the top-$K$ accuracy as the highest pose accuracy achieved among all $K^4$ hypotheses.
Fig.~\ref{fig:top_k} shows 
the curve of top-$K$ accuracies of our pose estimation network across all object instances, in terms of the mPCK\footnote{Please refer to Sec.~\ref{sec:exp} for more details on the mPCK metric.} metric on YCB-Video benchmark~\cite{xiang2017posecnn}.
We observe that pose estimation performance significantly improves when we initially increase $K$ from $1$ to $2$ and almost saturates at $K=4.$
This suggests that the inferred confidence score is ambiguous in only a small range, which makes sense especially for objects that have symmetric geometry or texture. 
The question is how we can resolve this ambiguity and further improve the pose estimation performance.
We now present a multi-view voting algorithm that selects the correct hypothesis from the top-$K$ hypothesis set.
% This is where we bring in the multi-view scenario where pose hypotheses in different views vote for each other so that the correct hypothesis can stand out.

\subsection{Hypothesis Voting}
\label{sec:hypo_voting}

To measure the difference between hypotheses from different views, we first transfer all hypotheses into view $1$ using the known camera poses of all $n$ views.
We consider a hypothesis set $\mathcal{H}=\{h_{1,1},\cdots,h_{i,j},\cdots,h_{n,K^4}\}$ from $n$ views, where $h_{i,j}$ indicates the pose hypothesis $j$ in view $i$ with respect to camera coordinate of view $1$.
% We denote $\mathcal{T}_{i}$ as the transformation for view $i$ so $\mathcal{T}_1$ is an identity transformation.
% Next, we pre-compute all pairwise distances $D(h_{i, j}, h_{p, q})$ for every two hypotheses  $h_{i, j}, h_{p, q} \in \mathcal{H}$.
To handle single-view ambiguity caused by symmetrical geometry, we test the consistency of ``fit'' to the observed data. More specifically, we employ the distance metric proposed by \cite{linemod-accv12} to measure the discrepancy between two hypothesis $h_1=(R_1, T_1)$ and $h_2=(R_2, T_2)$:
\begin{equation}
D(h_1, h_2) = \frac{1}{m}\sum_{x_1\in \mathcal{M}} \min_{x_2\in\mathcal{M}} \| (R_1x_1 + T_1) - (R_2x_2 + T_2) \|_2
\label{eq:dist_metric}
\end{equation}
where $\mathcal{M}$ denotes the set of 3D model points and $m=|\mathcal{M}|$.
$D(h_1, h_2)$ yields small distance when 3D object occupancies under poses $h_1$ and $h_2$ are similar, even if $h_1$ and $h_2$ have large geodesic distance on SO(3).
Finally, the voting score $V(h_{i, j})$ for $h_{i, j}$ is calculated as:
\begin{equation}
V(h_{i, j}) = \sum_{h_{p, q} \in \mathcal{H} \setminus h_{i, j}} \max\Big(\sigma - D(h_{i, j}, h_{p, q}),  0\Big)
\label{eq:voting}
\end{equation}
where $\sigma$ is the threshold for outlier rejection.
We select the hypothesis with the highest vote score as the final prediction.
Fig.~\ref{fig:mvmc_frame} (d) illustrates this multi-view voting process.

% One alternative way to combine multiple hypotheses is to simply ``max'' or ``average'' pool them.
% For ``max'' pooling, we compute the final estimate by using the bin and delta with highest confidence score, which fails to the exploit the top-K results and is essentially the same as in single-view case.
% In addition, the ``average'' of multiple SE(3) estimates can be performed by principled approaches such as ~\cite{chirikjian2018pose}.
% However, it is sensitive to incorrect predictions or outliers. 

\noindent
\textbf{Efficient Implementation.}
% In principle, the above hypothesis voting algorithm is capable of fusing arbitrary number of views.
% However, it becomes computationally expensive when the number of views is large and the time complexity of calculating pairwise distance $D(\cdot, \cdot)$ is high.
The above hypothesis voting algorithm is computationally expensive because the time complexity of Eq.~\ref{eq:dist_metric} is at least $O(m\log m)$ via a KDTree implementation.
% when the time complexity of calculating pairwise distance $D(\cdot, \cdot)$ is high.
% The time complexity of Eq.~\ref{eq:dist_metric} is at least $O(m\log m)$ using KDTree, which is still very inefficient.
Our solution is to decouple translation and rotation components in Eq.~\ref{eq:dist_metric} and approximate $D(h_1, h_2)$ by $\widetilde{D}(h_1, h_2)$:
\begin{equation}
\widetilde{D}(h_1, h_2) = \|T_1 - T_2\|_2 + \frac{1}{m}\sum_{x_1\in \mathcal{M}} \min_{x_2\in\mathcal{M}} \| R_1x_1 - R_2x_2 \|_2
\label{eq:approx_dist_metric}
\end{equation}
In fact, $\widetilde{D}(h_1, h_2)$ is an upper bound on $D(h_1, h_2)$: $D(h_1, h_2)\leq\widetilde{D}(h_1, h_2)$ for any $h_1$ and $h_2$,  because $\| (R_1x_1 + T_1) - (R_2x_2 + T_2) \|_2 \leq \|R_1x_1 - R_2x_2\| + \|T_1 - T_2\|$ based on the triangle inequality. 
Since the complexity of $\|T_1 - T_2\|$ is $O(1)$, we can focus on speeding up the computation of rotation distance $\frac{1}{m}\sum_{x_1\in \mathcal{M}} \min_{x_2\in\mathcal{M}} \| R_1x_1 - R_2x_2 \|_2$.
Our approach is to pre-compute a table of all pairwise distances between every two rotations from $N$ uniformly sampled rotation bins $\{\hat{R}_1,...,\hat{R}_N\}$ by \cite{yan2012almost}.
% $\{\hat{R}_1,...,\hat{R}_N\}$, computed by the same uniform sampling technique as used in Sec.~\ref{sec:bd_so(3)}, forms a uniform and dense coverage over SO(3).
For arbitrary $R_1$ and $R_2$, we search for their nearest neighbors $\hat{R}_{N_1(R_1)}$ and $\hat{R}_{N_1(R_2)}$ from $\{\hat{R}_1,...,\hat{R}_N\}$.
In turn, we approximate the rotation distance as follows:
\begin{equation}
\frac{1}{m}\sum_{x_1\in \mathcal{M}} \min_{x_2\in\mathcal{M}} \| R_1x_1 - R_2x_2 \|_2 \approx \frac{1}{m}\sum_{x_1\in \mathcal{M}} \min_{x_2\in\mathcal{M}} \| \hat{R}_{N_1(R_1)}x_1 - \hat{R}_{N_1(R_2)}x_2 \|_2
\label{eq:rot_approx}
\end{equation}
where the right hand side can be directly retrieved from the pre-computed distance table during inference.
When $N$ is large enough, the approximation error of Eq.~\ref{eq:rot_approx} has little effect on our voting algorithm.
% During inference, we can retrieve the error based on the nearest neighbors of $R_1$ and $R_2$.
In practice, we find the performance gain saturates when $N\geq1000$. 
Thus, the complexity of Eq.~\ref{eq:rot_approx} is $O(\log N)$ for nearest neighbor search, which is significantly smaller than $O(m\log m)$ of Eq.~\ref{eq:voting} ($m>>N$ in general). 
% As a consequence, this efficient algorithm is capable of significantly speeding up the multi-view voting scheme.

\section{Experiments}
\label{sec:exp}
In this section, we empirically evaluate our method on three large-scale datasets: YCB-Video~\cite{xiang2017posecnn}, JHUScene-50~\cite{chi-icra16} for 6-DoF pose estimation, and 
ObjectNet-3D~\cite{xiang2016objectnet3d} for viewpoint estimation.
Further, we conduct an ablative study to validate our three innovations for the single-view pose network.
% Finally, we qualitatively evaluate our method. 

% Implementation Details and Names of our Methods
\textbf{Evaluation Metric.}
For 6-DoF pose estimation, we follow the recently proposed metric ``ADD-S'' \cite{xiang2017posecnn}. 
The traditional metric~\cite{linemod-accv12} considers a pose estimate $h$ to be correct if $D(h, h^*)$ in Eq.~\ref{eq:dist_metric} is below a threshold with respect to the ground truth value $h^*$. 
``ADD-S'' improves this threshold-based metric by computing the area under the curve of the accuracy-threshold over different thresholds within a range (i.e. $[0, 0.1]$). 
We rename ``ADD-S'' as ``mPCK'' because it is essentially the mean of PCK accuracy~\cite{yang2011articulated}.  
For viewpoint estimation, we use Average Viewpoint Precision (AVP) used in PASCAL3D+~\cite{xiang2014beyond} and Average
Orientation Similarity (AOS) used in KITTI~\cite{geiger2012we}.

\textbf{Implementation Details.}
The number of nearest neighbors we use for soft binning is $4$ for SO(3) and $3$ for each of XYZ axes. 
We set binning scores as $\theta_1=\theta'_1=0.7$ and $\theta_2=\theta'_2=0.1$.
The number of rotation bins is $60$.
For XYZ binning, we use $10$ bins and $[s_{min}, s_{max}]=[-0.2, 0.2]$ for each axis when RGB-D data is used.
For inference on RGB data, we use $20$ bins, $[s_{min}, s_{max}]=[0.2, 0.8]$ for XY axes and $40$ bins, $[s_{min}, s_{max}]=[0.5, 4.0]$ for Z axis. 
In multi-view voting, we set the distance threshold $\sigma=0.02$ and the precomputed size of distance table as $2700$.
The input image to our single-view pose network is $64$x$64$.
The tiled class map is inserted at convolutional layer 15 with size $H=W=16$.
We use stochastic gradient descent with momentum $0.9$ to train our network from scratch.
The learning rate starts at 0.01 and decreases by one-tenth every $70000$ steps.
The batch size is $105$ for YCB-Video and $100$ for both JHUScene-50 and ObjectNet-3D.
We construct each batch by mixing equal number of data from each class.
We name our Multi-Class pose Network as ``MCN''. 
The multi-view framework using n views is called as ``MVn-MCN''.
Since MCN also infers instance mask, we use it to extract object point clouds when depth data is available and then run ICP to refined estimated poses by registering the object mesh to extracted object clouds.
We denote this ICP-based approach as ``poseCNN+ICP''. 

\subsection{YCB-Video}
\begin{table}[t]
\small
\centering
\renewcommand{\arraystretch}{0.9}% Tighter
\resizebox{0.95\textwidth}{!}{
\begin{tabular}{|  >{\centering\arraybackslash}m{1.05in} |  >{\centering\arraybackslash}m{0.4in} |  >{\centering\arraybackslash}m{0.4in} | >{\centering\arraybackslash}m{0.4in} || >{\centering\arraybackslash}m{0.4in} | >{\centering\arraybackslash}m{0.45in} |  >{\centering\arraybackslash}m{0.4in} | >{\centering\arraybackslash}m{0.4in} |  >{\centering\arraybackslash}m{0.4in} | >{\centering\arraybackslash}m{0.4in} |}
\hline
\multirow{2}{*}{Object}	&	\multicolumn{3}{c||}{RGB}  & \multicolumn{5}{c|}{RGB-D} \\
\hhline{|~|-|-|-|-|-|-|-|-|} & P-CNN ~\cite{xiang2017posecnn}  & MCN & MV5-MCN & 3D Reg. ~\cite{xiang2017posecnn} & P-CNN + ICP~\cite{xiang2017posecnn} & MCN &  MCN + ICP & MV5-MCN \\
\hline
002\_master\_chef\_can 	& 84.4 & 87.8 & \textbf{90.6} & 90.1 & 95.7 & 89.4 & 96.0 & \textbf{96.2} \\
\hline
003\_cracker\_box 		& \textbf{80.8} & 64.3 & 72.0 & 77.4 & \textbf{94.8} & 85.4 & 88.7 & 90.9 \\
\hline
004\_sugar\_box 		& 77.5 & 82.4 & \textbf{87.4} & 93.3 & \textbf{97.9} & 92.7 & 97.3 & 95.3 \\
\hline
005\_tomato\_can 		& 85.3 & 87.9 & \textbf{91.8} & 92.1 & 95.0 & 93.2 & 96.5 & \textbf{97.5} \\
\hline
006\_mustard\_bottle 	& 90.2 & 92.5 & \textbf{94.3} & 91.1 & \textbf{98.2} & 96.7 & 97.7 & 97.0 \\
\hline
007\_tuna\_fish\_can 	& 81.8 & 84.7 & \textbf{89.6} & 86.9 & 96.2 & 95.1 & \textbf{97.6} & 95.1 \\
\hline
008\_pudding\_box 		& \textbf{86.6} & 51.0 & 51.7 & 89.3 & \textbf{98.1} & 91.6 & 86.2 & 94.5 \\
\hline
009\_gelatin\_can 		& 86.7 & 86.4 & \textbf{88.5} & 97.2 & \textbf{98.9} & 94.6 & 97.6 & 96.0 \\
\hline
010\_potted\_meat\_can 	& 78.8 & 83.1 & \textbf{90.3} & 84.0 & 91.6 & 91.7 & 90.8 & \textbf{96.7} \\
\hline
011\_banana 			& 80.8 & 79.1 & \textbf{85.0} & 77.3 & 96.5 & 93.8 & \textbf{97.5} & 94.4 \\
\hline
019\_pitcher\_base 		& 81.0 & 84.8 & \textbf{86.1} & 83.8 & \textbf{97.4} & 93.8 & 96.6 & 96.2 \\
\hline
021\_bleach\_cleanser 	& 75.7 & 76.0 & \textbf{81.0} & 89.2 & 96.3 & 92.9 & \textbf{96.4} & 95.4 \\
\hline
024\_bowl 				& 74.2 & 76.1 & \textbf{80.2} & 67.4 & \textbf{91.7} & 82.6 & 76.0 & 82.0 \\
\hline
025\_mug 				& 70.0 & 91.4 & \textbf{93.1} & 85.3 & 94.2 & 95.3 & \textbf{97.3} & 96.8 \\
\hline
035\_power\_drill 		& 73.9 & 76.0 & \textbf{81.1} & 89.4 & \textbf{98.0} & 88.2 & 95.9 & 93.1 \\
\hline
036\_wood\_block 		& \textbf{63.9} & 54.0 & 58.4 & 76.7 & 93.1 & 81.5 & 93.5 & \textbf{93.6} \\
\hline
037\_scissors 			& 57.8 & 71.6 & \textbf{82.7} & 82.8 & \textbf{94.6} & 87.3 & 79.2 & 94.2 \\
\hline
040\_large\_marker 		& 56.2 & 60.1 & \textbf{66.3} & 82.8 & 97.8 & 90.2 & \textbf{98.0} & 95.4 \\
\hline
051\_large\_clamp 		& 34.3 & 66.8 & \textbf{77.5} & 67.6 & 81.5 & 91.5 & \textbf{94.0} & 93.3 \\
\hline
052\_larger\_clamp 		& 38.6 & 61.1 & \textbf{68.0} & 49.0 & 51.6 & 88.0 & 90.7 & \textbf{90.9} \\
\hline
061\_foam\_brick 		& \textbf{82.0} & 60.9 & 67.7 & 82.4 & 96.4 & 93.2 & \textbf{96.5} & 95.9 \\
\hhline{|=|=|=|=||=|=|=|=|=|}
All 					& 73.4 & 75.1 & \textbf{80.2} & 83.7 & 93.1 & 90.6 & 93.3 & \textbf{94.3} \\
\hline
\end{tabular}}
\caption{mPCK accuracies achieved by different methods on YCB-Video dataset~\cite{xiang2017posecnn}. The last row indicates the average-per-instance of mPCKs of all instances.}
\label{tab:posecnn}
\end{table}

% \begin{table}[t]
% \small
% \centering
% \begin{tabular}{|  >{\centering\arraybackslash}m{0.4in} |  >{\centering\arraybackslash}m{0.7in} |  >{\centering\arraybackslash}m{0.6in} | >{\centering\arraybackslash}m{0.6in} ||  >{\centering\arraybackslash}m{0.6in} |  >{\centering\arraybackslash}m{0.6in} | >{\centering\arraybackslash}m{0.6in} |  >{\centering\arraybackslash}m{0.6in} | >{\centering\arraybackslash}m{0.6in} |}
% \hline
% \multirow{2}{*}{YCB}	&	\multicolumn{3}{c||}{RGB}  & \multicolumn{4}{c|}{RGB-D} \\
% \hhline{|~|-|-|-|-|-|-|-|} & PoseCNN[30]  & MCN & MV5-MCN &  PoseCNN + ICP[30] & MCN &  MCN + ICP & MV5-MCN \\
% \hline
% All 					& 73.4 & 75.1 & \textbf{80.2}  & 93.1 & 90.6 & 93.3 & \textbf{94.3} \\
% \hline
% \end{tabular}
% \label{tab:posecnn_1}
% \end{table}

YCB-Video dataset~\cite{xiang2017posecnn} contains 92 real video sequences for 21 object instances.
80 videos along with 80,000 synthetic images are used for training and 2949 key frames are extracted from the remaining 12 videos for testing.
% The total number of images is 133,827 which is two full orders of magnitude larger than LINEMOD dataset~\cite{linemod-accv12}.
We fine tune the current state-of-the-art ``mask-RCNN''~\cite{he2017mask} on the training set as the detection system.
Following the same scenario in \cite{xiang2017posecnn}, we assume that one object appears at most once in a scene.
Therefore, we compute the bounding box of a particular object by finding the one with highest detection score of that object.  
For our multi-view system, one view is coupled with $5$ other randomly sampled views in the same sequence.
Each view outputs top-$3$ results from each space of SO(3), X, Y and Z and in turn $3^4=81$ pose hypotheses. 
% Therefore, the number of pose hypotheses from each single view is .

Table~\ref{tab:posecnn} reports mPCK accuracies of our methods and variants of poseCNN~\cite{xiang2017posecnn} (denoted as ``P-CNN''). 
All methods are trained and tested following the same experiment setting defined in \cite{xiang2017posecnn}.
We first observe that the multi-view framework (MV5-MCN) consistently improves the single-view network (MCN) across different instances and achieves the overall state-of-the-art performance.
Such improvement is more significant on RGB data, where the mPCK margin between MV5-MCN and MCN is $5.1\%$ which is much larger than the margin of $1.0\%$ on RGB-D data for all instances.
This is mainly because single-view ambiguity is more severe without depth data.  
Subsequently, MCN outperforms poseCNN by $1.7\%$ on RGB and MCN+ICP is marginally better than poseCNN+ICP by $0.2\%$ on RGB-D.
We can see that MCN achieves more balanced performance than poseCNN across different instances. 
% That is, poseCNN achieves top mPCK accuracies of some individual classes but its bad result is much worse than MCN.
For example, poseCNN+ICP only obtains $51.6\%$ on class ``052\_larger\_clamp'' which is $24.4\%$ lower than the minimum accuracy of a single class by MCN+ICP.
This can be mainly attributed to our class fusion design in learning discriminative class-specific feature so that similar objects can be well-separated in feature space (e.g. ``051\_large\_clamp'' and ``052\_larger\_clamp''). 
We also observe that MCN is much inferior to PoseCNN on some instances such as foam brick.
This is mainly caused by larger detection errors (less than 0.5 IoU with ground truth) on these instances.

We also run MCN over ground truth bounding boxes and the overall mPCKs are $86.9\%$ on RGB ($11.8\%$ higher than the mPCK on detected bounding boxes) and $91.0\%$ on RGB-D ($0.4\%$ higher the mPCK on detected bounding boxes).
This indicates that MCN is sensitive to detection error on RGB while being robust on RGB-D data. 
The reason is that we rely on the image scale of bounding box to recover 3D translation for RGB input.
In addition, we obtain high instance segmentation accuracy\footnote{The ratio of the number of pixels with correctly predicted mask label versus all} of MCN across all object instances: $89.9\%$ on RGB and $90.9\%$ on RGB-D.
% Although mask label is only an auxiliary concept for deep supervision, MCN achieves high average of mask estimation accuracies : 
This implies that MCN does actually learn the intermediate foreground mask as part of pose prediction. 
We refer readers for more numerical results in supplementary material, including segmentation accuracies, PCK curves of MCN and mPCK accuracies on groundtruth bounding box on individual instance.
Last, we show some qualitative results in upper part of Fig.~\ref{fig:qual_result}.
We can see that MCN is capable of predicting object pose under occlusion and MV5-MCN further refines the MCN result.

\subsection{JHUScene-50}
\label{sec:exp_scene50}

\begin{table}[t]
\small
\centering
\renewcommand{\arraystretch}{0.9}% Tighter
\resizebox{0.9\textwidth}{!}{
\begin{tabular}{|  >{\centering\arraybackslash}m{0.55in} |  >{\centering\arraybackslash}m{0.65in} |  >{\centering\arraybackslash}m{0.5in}  | >{\centering\arraybackslash}m{0.6in}  || >{\centering\arraybackslash}m{0.65in} | >{\centering\arraybackslash}m{0.65in} |  >{\centering\arraybackslash}m{0.5in}  | >{\centering\arraybackslash}m{0.6in}  | }
\hline
\multirow{2}{*}{Object}	&	\multicolumn{3}{c||}{RGB}  & \multicolumn{4}{c|}{RGB-D} \\
\hhline{|~|-|-|-|-|-|-|-|} & Manifold~\cite{balntas2017pose} & MCN & MV5-MCN & ObjRec.~\cite{objransac-accv10} & Manifold~\cite{balntas2017pose} & MCN  & MV5-MCN \\
\hline
drill\_1 		& 10.6 & 33.4 & \textbf{36.5} & 14.5 & 70.3 & 76.8 & \textbf{78.1} \\
\hline
drill\_2 		& 9.9 & 48.8 & \textbf{54.5} & 2.9 & 49.0 & 76.6 & \textbf{80.1} \\
\hline
drill\_3 		& 7.6 & 45.5 & \textbf{48.0} & 3.7 & 50.9 & 81.5 & \textbf{85.4} \\
\hline
drill\_4 		& 9.3 & 41.6 & \textbf{45.5} & 6.5 & 51.4 & 82.0 & \textbf{87.1} \\
\hline
hammer\_1 		& 5.0 & 24.9 & \textbf{30.2} & 8.1 & 38.7 & 80.1 & \textbf{87.6} \\
\hline
hammer\_2 		& 5.1 & 28.3 & \textbf{33.4} & 10.7 & 35.5 & 81.2 & \textbf{91.5} \\
\hline
hammer\_3 		& 7.8 & 26.2 & \textbf{31.2} & 8.6 & 47.8 & 83.1 & \textbf{88.1} \\
\hline
hammer\_4 		& 5.1 & 17.2 & \textbf{20.6} & 3.8 & 38.3 & 73.8 & \textbf{87.8} \\
\hline
hammer\_5 		& 5.2 & 37.1 & \textbf{44.4} & 9.6 & 35.0 & 78.0 & \textbf{86.3} \\
\hline
sander 			& 10.7 & 35.6 & \textbf{39.5} & 9.5 & 54.3 & \textbf{76.0} & 75.5 \\
\hhline{|=|=|=|=||=|=|=|=|}
All 			& 7.6 & 33.9 & \textbf{38.4} & 7.8 & 47.1 & 78.9 & \textbf{84.8} \\
\hline
\end{tabular}}
\caption{mPCK accuracies of all objects in JHUScene-50 dataset~\cite{chi-icra16}. The last row indicates the average-per-class of mPCKs of all object instances. Best results are highlighted in bold.}
\label{tab:jhu}
\end{table}

% \begin{table}[t]
% \small
% \centering
% \begin{tabular}{|  >{\centering\arraybackslash}m{0.55in} |  >{\centering\arraybackslash}m{0.65in} |  >{\centering\arraybackslash}m{0.5in}  | >{\centering\arraybackslash}m{0.6in}  || >{\centering\arraybackslash}m{0.65in} | >{\centering\arraybackslash}m{0.65in} |  >{\centering\arraybackslash}m{0.5in}  | >{\centering\arraybackslash}m{0.5in}  | }
% \hline
% \multirow{2}{*}{JHU}	&	\multicolumn{3}{c||}{RGB}  & \multicolumn{4}{c|}{RGB-D} \\
% \hhline{|~|-|-|-|-|-|-|-|} & Pose Manifold~\cite{balntas2017pose} & MCN & MV5-MCN & ObjRec.~\cite{objransac-accv10} & Pose Manifold~\cite{balntas2017pose} & MCN  & MV5-MCN \\
% \hline
% All 			& 7.6 & 33.9 & \textbf{38.4} & 7.8 & 47.1 & 78.9 & \textbf{84.8} \\
% \hline
% \end{tabular}
% \label{tab:jhu_1}
% \end{table}

JHUScene-50~\cite{chi-icra16} contains 50 scenes with diverse background clutter and severe object occlusion.
Moreover, the target object set consists of $10$ hand tool instances with similar appearance.
% with heavier occlusion and more diverse background clutter than YCB-Video and LINEMOD~\cite{linemod-accv12} datasets. 
% Moreover, many scenes contain objects with similar texture and 3D geometry.
% There are $5000$ labeled frames with $22520$ pose annotations in total.
Only textured CAD models are available during training and all $5000$ real image frames comprise the test set.
To cope with our pose learning framework, we simulate a large amount of synthetic data by rendering densely cluttered scenes similar to the test data, where objects are randomly piled on a table.
We use UnrealCV~\cite{qiu2017unrealcv} as the rendering tool and generate 100k training images. 

\begin{figure}[t]
  \centering
    \includegraphics[width=0.95\linewidth]{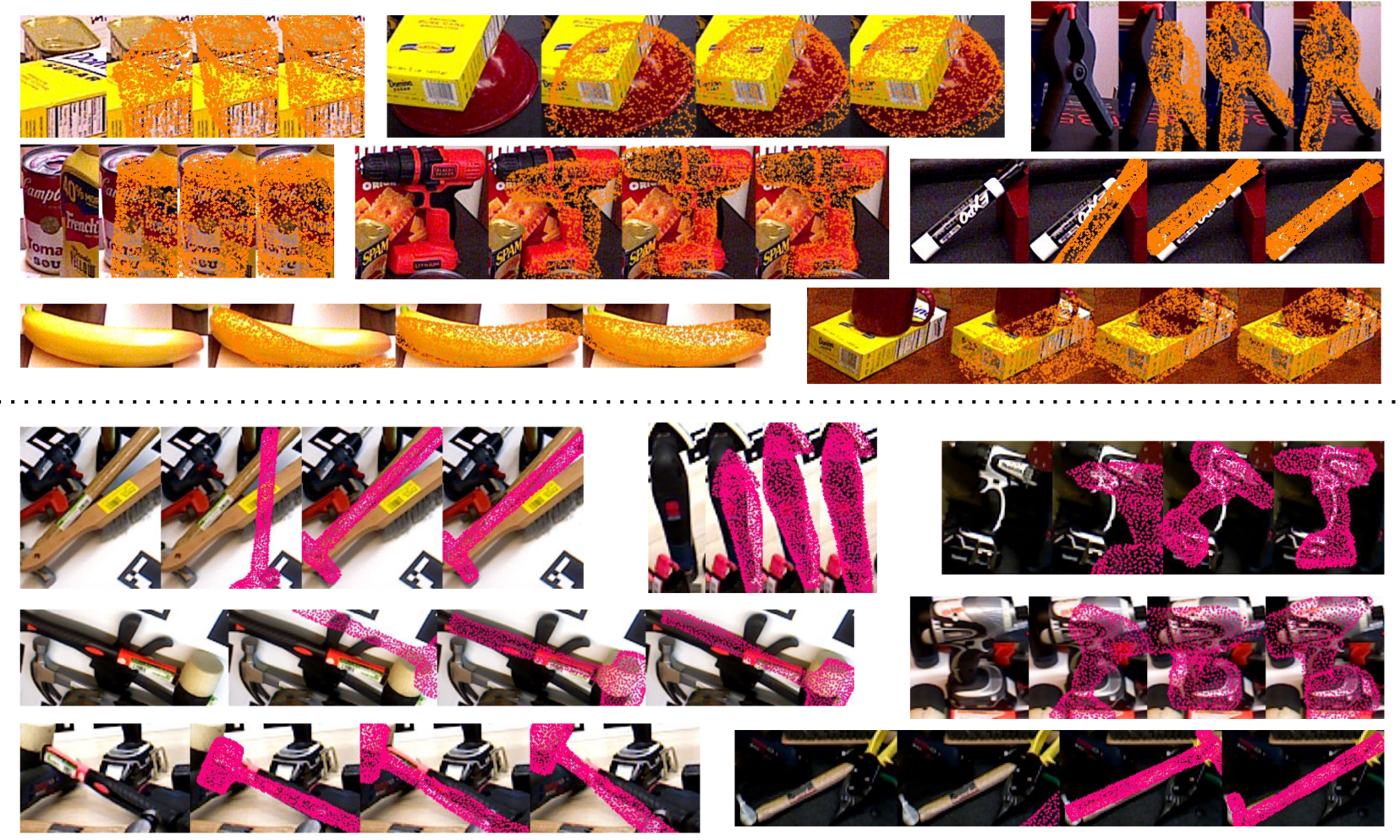}
  \caption{Illustration of pose estimation results by MCN on YCB-Video (upper) and JHUScene-50 (bottom). The projected object mesh points that are transformed by pose estimates are highlighted by orange (YCB-Video) and pink (JHUScene-50). From left to right of each data, we show original ROI, MCN estimates on RGB, MCN estimates on RGB-D and MV5-MCN estimates on RGB-D.}
 \label{fig:qual_result}
\end{figure}

We compare MCN and MV5-MCN with the baseline method ObjRecRANSAC\footnote{https://github.com/tum-mvp/ObjRecRANSAC}~\cite{objransac-accv10} in JHUScene-50 and one recent state-of-the-art pose manifold learning technique~\cite{balntas2017pose}\footnote{We re-implement this method because the source code is not publicly available.}.
All methods are trained on the same synthetic training set and tested on the $5000$ real image frames from JHUScene-50.
We compute 3D translation for ~\cite{balntas2017pose} by following the same procedure used in \cite{linemod-accv12}.
We evaluate different methods on the ground truth locations of all objects.
Table~\ref{tab:jhu} reports mPCK accuracies of all methods.
We can see that MCN significantly outperforms other comparative methods by a large margin, though MCN performs much worse than on YCB-Video mainly because of the severe occlusion and diverse cluttered background in JHUScene-50.
Additionally, we observe that MV5-MCN is superior to MCN on both RGB and RGB-D data.
The performance gain on RGB-D data achieved by MV5-MCN is much larger than the one on YCB-Video, especially for the hammer category due to the symmetrical 3D geometry. 
We visualize some results of MCN and MV5-MCN in the bottom of Fig.~\ref{fig:qual_result}.
The bottom-right example shows MV5-MCN corrects the orientation of MCN result which frequently occurs for hammer.

\subsection{ObjectNet-3D}

\begin{table}[t]
\small
\centering
\resizebox{0.9\textwidth}{!}{
\begin{tabular}{|  >{\centering\arraybackslash}m{0.5in} | >{\centering\arraybackslash}m{0.9in} || >{\centering\arraybackslash}m{0.95in} |  >{\centering\arraybackslash}m{0.55in}  || >{\centering\arraybackslash}m{0.95in}  | >{\centering\arraybackslash}m{0.55in}  | }
\hline
\multirow{2}{*}{}	 & mAP 	&	\multicolumn{2}{c||}{AOS}  & \multicolumn{2}{c|}{AVP} \\
\hhline{|~|-|-|-|-|-|} & Fast R-CNN~\cite{girshick2015fast} & ObjectNet-3D~\cite{xiang2016objectnet3d} & MCN & ObjectNet-3D~\cite{xiang2016objectnet3d} & MCN \\
\hline
Accuracy 		&	61.6  & 51.9 & \textbf{56.0} & 39.4 (64.0)  & \textbf{50.0} (\textbf{81.2})\\
\hline 
\end{tabular}}
\caption{Accuracies of object pose estimation on ObjectNet-3D benchmark~\cite{xiang2016objectnet3d}. All methods perform over the same set of detected bounding boxes estimated by Fast R-CNN~\cite{girshick2015fast}. Best results on both AOS and AVP metrics are shown in bold. For AVP, we also report $\frac{\text{AVP}}{\text{mAP}}$ in parentheses.}
\label{tab:obj3d}
\end{table}

To evaluate the scalability of our method, we conduct an experiment on ObjectNet-3D which consists of viewpoint annotation of $201,888$ instances from $100$ object categories.
In contrast to most existing benchmarks~\cite{xiang2017posecnn,chi-icra16,linemod-accv12} which target indoor scenes and small objects, ObjectNet-3D covers a wide range of outdoor environments and diverse object categories such as airplane.
We modify the MCN model by only using the rotation branch for viewpoint estimation and removing the deep supervision of object mask because the object mask is not available in ObjectNet-3D.
To our knowledge, only \cite{xiang2016objectnet3d} reports viewpoint estimation accuracy on this dataset, where a viewpoint regression branch is added along with bounding box regression in the Fast R-CNN architecture~\cite{girshick2015fast}.
For the fair comparison, we use the same detection results for \cite{xiang2016objectnet3d} as the input to MCN.
Because ObjectNet-3D only provides detection results on the validation set, we train our model on the training split and test on the validation set. 
Table~\ref{tab:obj3d} reports the viewpoint estimation accuracies of different methods on the validation set,in terms of two different metrics AVP~\cite{xiang2014beyond} and AOS~\cite{geiger2012we}.
The detection performance in mAP is the upperbound of AVP. 
The numbers in parentheses are the ratios of AVP versus mAP.
We can see that MCN is significantly superior to the large-scale model~\cite{xiang2016objectnet3d} on both AOS and AVP, even if \cite{xiang2016objectnet3d} actually optimizes the network hyper-parameters on the validation set.
This shows that MCN can be scaled to a large-scale pose estimation problem.
Moreover, object instances have little overlap between training and validation sets in ObjectNet-3D, which indicates that MCN can generalize to unseen object instances within a category. 

\subsection{Ablative Study}

\begin{table}[t]
\small
\centering
\resizebox{0.9\textwidth}{!}{
\begin{tabular}{|  >{\centering\arraybackslash}m{1.4in} |  >{\centering\arraybackslash}m{0.6in} |  >{\centering\arraybackslash}m{0.5in}  | >{\centering\arraybackslash}m{0.8in}  || >{\centering\arraybackslash}m{0.7in}  | >{\centering\arraybackslash}m{0.5in} | }
\hline
\multirow{2}{*}{Method}	&	\multicolumn{3}{c||}{RGB}  & \multicolumn{2}{c|}{RGB-D} \\
\hhline{|~|-|-|-|-|-|} & YCB-Video & JHU & ObjectNet-3D & YCB-Video & JHU \\
\hline
plain								& 61.0 & 25.0 & 51.7 / 38.3  & 61.8 & 19.6 \\
\hline
BD + Seg 							& 66.2 & 26.3 & 50.3* / 41.3*  & 89.5 & 70.0 \\
\hline
BD + TC 							& 68.5 & 29.3 & \textbf{56.0} / \textbf{50.0} & 90.1 & 76.4 \\
\hline
Sep-Branch + Seg + BD 				& 73.8 & 31.6 & 52.5* / 42.9* & 90.2 & 77.7 \\
\hline
Sep-Net + Seg + BD					& 62.1 & 28.7 & NA & 87.1 & 66.9 \\
\hhline{|=|=|=|=|=|=|}
MCN (Seg + TC + BD) 				& \textbf{80.2} & \textbf{33.9} & NA & \textbf{90.8} & \textbf{78.9} \\
\hline
\end{tabular}}
\caption{An ablative study of different variants of pose estimation architectures on YCB-Video, JHUScene-50 and ObjectNet-3D. We follow the same metrics as we evaluate in previous sections. For ObjectNet-3D, we report accuracies formatted as AOS / AVP. The ``*'' symbol indicates that no segmentation mask is used in training because it is unavailable in ObjectNet-3D.}
\label{tab:abl_study}
\end{table}

In this section, we empirically validate the three innovations introduced in MCN: bin \& delta representation (``BD''), tiled class map (``TC'') and deep supervision of object segmentation (``Seg'').
Additionally, we also inspect the baseline architectures: separate network for each object (``Sep-Net'') and separate output branch for each object (``Sep-Branch''), as shown in Fig.~\ref{fig:mvmc_frame} (a) and Fig.~\ref{fig:mvmc_frame} (b) respectively.
To remove the effect of using ``BD'', we directly regress quaternion and translation (plain) as the comparison.
Table~\ref{tab:abl_study} presents accuracies of different methods on all three benchmarks.
We follow previous sections to report mPCK for YCB-Video and JHUScene-50, and AOS/AVP for ObjectNet-3D.
Because ObjectNet-3D does not provide segmentation groundtruth, we remove module ``Seg'' in all analysis related to ObjectNet-3D.
Also, we do not report accuracy of ``Sep-Net'' on ObjectNet-3D because it requires 100 GPUs for training.
We have three main observations: 1. When removing any of the three innovations, pose estimation performance consistently decreases. Typically, ``BD'' is a more critical design than ``Seg'' and tiled class map because the removal of BD causes larger performance drop; 2. ``Sep-Branch'' coupled with ``BD'' and ``Seg'' appears to be the second best architecture, but it is still inferior to MCN especially on YCB-Video and ObjectNet-3D. Moreover, the model size of ``Sep-Branch'' grows rapidly with the increasing number of classes; 3. ``Sep-Net'' is expensive in training and it performs substantially worse than MCN because MCN exploits diverse data from different classes to reduce overfitting. 

\section{Conclusion}
\label{sec:conclude}
We present a unified architecture for inferring 6-DoF object pose from single and multiple views. 
We first introduce a single-view pose estimation network with three innovations: a new bin \& delta pose representation, the fusion of tiled class map into convolutional layers and deep supervision of object mask at intermediate layer.
These modules enable a scalable pose learning architecture for large-scale object classes and unconstrained background clutter.
Subsequently, we formulate a new multi-view framework for selecting single-view pose hypotheses while considering ambiguity caused by object symmetry.
In the future, an intriguing direction is to embed the multi-view procedure into the training process to jointly optimize both single-view and multi-view performance.
Also, the multi-view algorithm can be improved to maintain a fixed number of ``good'' hypotheses for any incremental update given a new frame.

{
\myparagraph{Acknowledgments} 
This work is supported by the IARPA DIVA program and the National Science Foundation under grants IIS-127228 and IIS-1637949.
}

\clearpage

\bibliographystyle{splncs04}
\bibliography{reference}
\end{document}